\documentclass{article} 
\usepackage{iclr2026_conference,times}

\usepackage{amsmath,amsfonts,bm}

\def\eqref#1{equation~\ref{#1}}

\def\1{\bm{1}}

\DeclareMathAlphabet{\mathsfit}{\encodingdefault}{\sfdefault}{m}{sl}
\SetMathAlphabet{\mathsfit}{bold}{\encodingdefault}{\sfdefault}{bx}{n}

\usepackage{hyperref}
\usepackage{url}
\usepackage{graphicx}
\usepackage{amsmath}
\usepackage{mathrsfs}
\usepackage{multirow} 
\usepackage[table]{xcolor}
\usepackage{pifont}
\usepackage{enumitem}

\bibpunct{(}{)}{,}{n}{,}{,}  
\let\cite\citep              

\usepackage{tikz}
\newcommand{\blackcircle}[1]{%
    \begin{tikzpicture}[baseline=-0.9ex]
        \node[circle,fill=black,text=white,inner sep=0.2pt,minimum size=1pt] {#1};
    \end{tikzpicture}%
}

\title{Protein as a Second Language for LLMs}

\author{Xinhui Chen\textsuperscript{\rm 1}\quad 
Zuchao Li\textsuperscript{\rm 1\thanks{Zuchao Li (zcli-charlie@whu.edu) is the corresponding author with School of Artificial Intelligence, Wuhan University.}}\quad 
Mengqi Gao\textsuperscript{\rm 1}\quad
Yufeng Zhang\textsuperscript{\rm 1}\quad
Chak Tou Leong\textsuperscript{\rm 2}\quad \\
\textbf{Haoyang Li}\textsuperscript{\rm 3}\quad 
\textbf{Jiaqi Chen}\textsuperscript{\rm 3,4} \\
\textsuperscript{\rm 1}Wuhan University\quad
\textsuperscript{\rm 2}Hong Kong Polytechnic University\quad
\textsuperscript{\rm 3}Stanford University\quad
\textsuperscript{\rm 4}Topify AI
}

\iclrfinalcopy 
\begin{document}

\maketitle

\begin{abstract}

Deciphering the function of unseen protein sequences is a fundamental challenge with broad scientific impact, yet most existing methods depend on task-specific adapters or large-scale supervised fine-tuning. We introduce the ``\textbf{\textit{Protein-as-Second-Language}}'' framework, which reformulates amino-acid sequences as sentences in a novel symbolic language that large language models can interpret through contextual exemplars. Our approach adaptively constructs sequence–question–answer triples that reveal functional cues in a zero-shot setting, without any further training. To support this process we curate a bilingual corpus of 79,926 protein–QA instances spanning attribute prediction, descriptive understanding, and extended reasoning. Empirically, our method delivers consistent gains across diverse open-source LLMs and GPT-4o, achieving up to 17.2\% ROUGE-L improvement (average +7\%) and even surpassing fine-tuned protein-specific language models. These results highlight that generic LLMs, when guided with protein-as-language cues, can outperform domain-specialized models, offering a scalable pathway for protein understanding in foundation models.
\end{abstract}

 \section{Introduction}

Proteins are indispensable molecular machines of life, driving key functions such as maintaining cell structure and enabling cell communication.
Their three-dimensional architectures, catalytic activities, interaction networks, and evolutionary trajectories are all encoded within a linear sequence composed of twenty amino-acid characters~\cite{kitadai2018origins,xiao2025protein}. 
Therefore, the core of understanding protein function lies in accurately ``reading'' and ``translating'' the biological meaning contained within these amino-acid sequences~\cite{clark2011analysis,koonin2002sequence}. 
However, this task is fraught with challenges. Although the amino acid sequence is formally like a language—possessing a fixed character set (over 20 genetically encoded amino acids) and potential grammatical rules (physicochemical laws)—the mapping relationship from the one-dimensional sequence to the three-dimensional structure and function is extremely complex and highly context-dependent~\cite{rost1998protein,wang2005context}.
Consequently, the central challenge of ``what cellular function does an unknown amino acid sequence encode?'' still lacks a comprehensive solution.

To address this challenge, research efforts on protein understanding can be broadly categorized into two dominant paradigms:
\textit{protein representation learning} and \textit{protein–language alignment modeling}.
Protein representation learning sees amino-acid sequences as a standalone modality like language and visual, acquires universal protein representations through self-supervised pre-training on large-scale amino-acid sequences, and then attaches lightweight decoders to predict structure or function~\cite{zhang2022ontoprotein,brandes2022proteinbert,lin2023evolutionary,su2023saprot,chen2024xtrimopglm,wu2024tcr,wu2024proteinclip}. While this paradigm excels in the universality of its embeddings and in mining deep sequential patterns, these embeddings still rely on additional ``interpreters'', \emph{i.e.}, post-processing adapters, to be converted into human-understandable explanations.
Protein–language alignment modeling, in contrast, co-trains on paired protein sequences and their textual descriptions, establishing a bidirectional mapping within a shared latent space that enables end-to-end text-based question answering~\cite{xu2022protranslator,pei2023biot5,guo2023proteinchat,abdine2024prot2text,wang2024protchatgpt,xiao2024proteingpt}. Although this route bypasses downstream adapters, it is intrinsically bound to large-scale paired data and often requires re-fine-tuning whenever the output format or downstream objective shifts.
In summary, both of these approaches face bottlenecks of large training data requirements, high computational costs, and limited generalization ability.

\paragraph{Protein as Second Language.} Reflecting on the human cognitive process, we observe that humans exhibit remarkable efficiency and generalization ability when learning a brand-new symbolic system (i.e., a new language). The key lies in their ability to rely on and transfer their existing native language knowledge system~\cite{gass2020second, jarvis2008crosslinguistic}.
Given the aforementioned ``linguistic'' properties of protein sequences—possessing a compositional structure and contextual semantics—and our goal of understanding their function using natural language, we propose a novel perspective: to treat protein sequences as a symbolic system that can be learned and interpreted by large language models (LLMs) as a ``second language''.

Analogous to how humans acquire a second language, \emph{i.e.}, by encountering new words in context and inferring their meaning and usage, we propose a protein language learning framework in which an LLM acquires protein semantics and reasoning ability through context-driven exposure that grounds sequence patterns in functional and structural examples. This framework adaptively constructs learning contexts for a given protein understanding goal, enabling rapid acquisition of target protein knowledge without additional training or sacrificing generalization.
To support effective learning, we constructed a ``bilingual'' dataset of 79,926 protein-sequence–question–answer triples covering functional, descriptive, and extended-information queries. 
Across Protein2Text~\cite{xu2023protst}, Mol-Instructions\cite{fang2023mol} and ProtDescribe-QA~\cite{Protein2Text2025}, our framework raises the average ROUGE-L by 7\% across diverse open-source models and GPT-4o, with a maximum gain of 17.2\%, without any task-specific fine-tuning.
Our contributions are as follows:
\begin{itemize}[leftmargin=0.6cm, itemsep=0.05cm]
\item We introduce the “\textbf{\textit{Protein-as-Second-Language}}” conceptual framework, which recasts amino-acid sequences as a second language that can be acquired via in-context learning, enabling efficient and generalized protein understanding.
\item We construct \textbf{\textit{a protein-natural language bilingual dataset}} that spans four task families: attribute-based QA, True or False QA, descriptive-text QA, and extended-information QA, to support effective protein language learning and benchmarking.
\item We present a protein language learning framework that adaptively constructs learning contexts for protein understanding, yielding significant gains for both open-source models and GPT-4o, enabling them to outperform domain-specialized models without additional training.
\end{itemize}

\section{Related Work}

\subsection{Language Models in Protein}

Protein representation learning with protein language models (PLMs) extends the Transformer to amino-acid strings, producing dense embeddings for property prediction~\cite{hayes2025simulating,brandes2022proteinbert,elnaggar2021prottrans,ProtGO,cao2021tale, chen2024xtrimopglm,chen2024unifying} or generative design~\cite{madani2023large, nijkamp2023progen2, lv2024prollama, ferruz2022protgpt2}. 
%
Because these models are trained exclusively on amino acid sequences, their outputs remain latent vectors that external classifiers must translate into human-readable function.
To obviate this indirection, protein–language alignment modeling has emerged, which jointly connects sequences with textual descriptions via (i) contrastive objectives mapping proteins and sentences into a shared space~\cite{protst, wu2024proteinclip}, (ii) bioknowledge-augmented pre-training on curated protein–text corpora~\cite{ferruz2022protgpt2, taylor2022galactica, lv2024prollama, BioT5, zhuo2024protllm, liu2024evollama}, or (iii) multi-modal LLMs that graft protein encoders onto frozen language backbones~\cite{liu-etal-2024-prott3, abdine2024prot2text, wang2024protchatgpt, chen2024unifying, ma2025prottex, xiang2024fapm}.
While effective, these approaches entail costly retraining or gradient updates and risk catastrophic forgetting when scaled to larger LLMs~\cite{kirkpatrick2017overcoming,wu2025rethinking}, prompting a shift toward parameter-efficient adaptation.

\subsection{Protein QA Datesets}
Datasets that couple proteins with natural-language annotations have become the empirical bedrock for developing protein–text hybrid systems. At present, two complementary families of corpora dominate the landscape. The first centers on protein captioning: given an amino-acid sequence alone, the objective is to generate a concise textual description. Representative instances include the richly annotated Swiss-Prot collection~\cite{bairoch2000swiss}, the ProteinKG resource~\cite{zhang2022ontoprotein} and ProtDescribe~\cite{xu2023protst}. The second family targets protein question answering: here, both a sequence and a natural-language query are supplied, and the model is required to synthesize an answer grounded in the provided protein. Curated examples span Mol-Instructions~\cite{fang2023mol}, UniProtQA~\cite{luo2024biomedgpt}, ProteinLMBench~\cite{shen2024fine}, VenusX~\cite{tan2025venusxunlockingfinegrainedfunctional} and Protein2Text-QA~\cite{Protein2Text2025}.

\section{Protein as Second Language}

We introduce ``Protein-as-Second-Language'', a framework that treats amino-acid sequences as a new symbolic system to be learned much like humans acquire a foreign language. Just as learners infer the meaning of unfamiliar words by repeatedly encountering them in context, we construct a \textbf{\textit{protein–natural language bilingual dataset}} (Sec.~\ref{sec:dataset}) and design an \textbf{\textit{adaptive context construction mechanism}} (Sec.~\ref{sec:framwork}) to provide such contextual exposure. In this way, our framework enables LLMs to acquire protein semantics through exemplars rather than through extensive re-training.

\subsection{Bilingual Dataset Construction}
\label{sec:dataset}

\begin{figure}[ht]
\centering
\includegraphics[width=\linewidth]{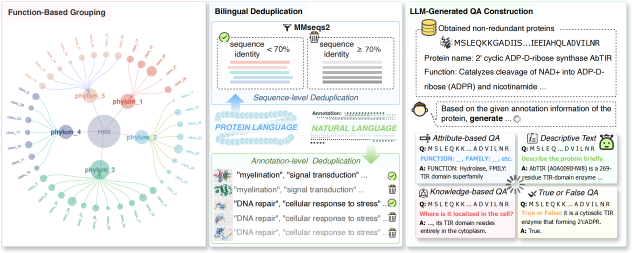}
\caption{\textbf{The overview of data construction of our bilingual protein–QA dataset.}}
\label{fig:pipline}
\end{figure}

We curate our bilingual dataset in three steps (Figure~\ref{fig:pipline}). Starting from 573,661 Swiss-Prot~\cite{bairoch2000swiss} entries with gene ontology (GO) annotations, we avoid directly converting all annotations, as this would introduce heavy redundancy; instead, we construct a balanced sample. Specifically, (i) we prune the GO-directed acyclic graph (GO-DAG) to obtain representative functional categories and group proteins accordingly (Sec.~\ref{sec:step1}), (ii) perform bilingual deduplication by clustering sequences within each protein group and sampling proteins with diverse functional annotation (Sec.~\ref{sec:step2}), and (iii) use DeepSeek-R1~\cite{guo2025deepseek} to generate attribute, knowledge, descriptive, and true/false QA pairs, yielding 79,926 high-quality protein–QA triples (Sec.~\ref{sec:step3}).

\subsubsection{Function-Based Grouping}
\label{sec:step1}

To enable representative sampling across functional categories, the dataset is partitioned according to the GO hierarchy. Directly using the raw directed acyclic graph (DAG) risks over-fragmentation from overly fine, sparsely populated terms, and excessive generalization near the root. To address this, we adapt a pruning strategy inspired by decision tree simplification~\cite{mondon1984classification}, where complexity is managed through a penalty to avoid overfitting. This strategy aims to retain an optimal set of GO terms as functional grouping nodes. It balances granularity and coverage, ensuring that the retained nodes represent biologically diverse yet statistically well-supported categories for downstream sampling.

The pruning process is driven by two main criteria:
(1) A node is retained if it meets the \textit{\textbf{minimum support threshold}}, which ensures that the node has a sufficient number of associated proteins, and does not exhibit significant child imbalance.
(2) If the \textit{\textbf{child-imbalance ratio}} is high, meaning the protein distribution among a node's child terms is uneven, the parent node is retained, even if the child nodes fail to meet the minimum support threshold.

\paragraph{Minimum Support Threshold}
A node is retained only if the number of associated proteins meets a depth-adjusted threshold $m(d)$, which adapts based on the node's depth in the GO hierarchy. The threshold is calculated as:
\begin{equation}
m(d) = \lambda \cdot C_{tot}\cdot(1+\beta d)
\end{equation}
where $C_{tot}$ is the total protein count, $d$ is the node depth, and $\lambda$ and $\beta$ are constants. This dynamic threshold is designed to prevents deep nodes from splitting infinitely due to overly small absolute values.

\paragraph{Child-Imbalance Ratio}
The child-imbalance ratio is applied to assess whether the child nodes of a given term are too imbalanced. The imbalance ratio $\rho(v)$ is computed as the ratio of the largest to the smallest protein count among the child nodes: 

\begin{equation}
p(u)=\frac{\displaystyle\max_{u\in C^{+}(v)} C(u)}
          {\displaystyle\min_{u\in C^{+}(v)} C(u)}
\end{equation}

where $C^{+}(v)$ represents the set of valid child nodes with non-zero protein counts. If the imbalance ratio $\rho(v)$ exceeds a specified threshold $\tau(d)$, the parent node $v$ is retained to preserve the biological diversity. This threshold is adjusted dynamically with the depth $d$ to allow for greater flexibility at deeper levels of the hierarchy:

\begin{equation}
\tau(d) = \tau_0 \cdot \alpha^d
\end{equation}
where $\tau_0$ is the base threshold, and $\alpha$ is a scaling factor.

By applying these two criteria, the pruning process is carried out recursively, allowing the algorithm to adaptively prune the GO DAG and identify the most relevant, biologically diverse functional groups.

\subsubsection{Bilingual deduplication }
\label{sec:step2}

After grouping by GO term, proteins within the same node often exhibit high similarity, as they represent homologous proteins. 
To address this, we use MMseqs2~\cite{steinegger2017mmseqs2} for sequence clustering within each GO node, applying a 70\% \textbf{\textit{amino acid sequence similarity}} threshold. From each cluster, a single representative sequence is selected. This threshold efficiently removes redundant sequences with minimal functional variation while preserving functional diversity.

While sequence similarity-based redundancy removal effectively reduces sequence-level redundancy, it does not necessarily capture functional divergence. Specifically, sequence similarity below 70\% does not imply functional divergence, and substantial functional redundancy may still exist within the set~\cite{devos2000practical}. To address this, we focus on \textbf{\textit{annotation semantic similarity}}, quantifying the functional relationships between proteins based on their GO annotations. Inspired by the simGIC method~\cite{pesquita2008metrics} for calculating GO terms semantic similarity, we calculate the Protein Functional Information Content $\mathrm{IC}_{\text{protein function}}$ for each protein, which is the sum of the Information Content (IC) of all associated GO terms and their ancestral terms. The IC of each GO term is calculated based on its frequency in the dataset, using the total protein set after sequence redundancy removal. The $\mathrm{IC}_{\text{protein function}}$ value for each Protein ID is computed as:

\begin{equation}
\text{IC}_{\text{protein function}} = \sum_{g \in \text{GO terms of } p} \text{IC}(g) + \sum_{g' \in \text{ancestors of GO terms of } p} \text{IC}(g').
\end{equation}

This provides a quantitative measure of each protein's functional information, capturing both direct and indirect annotations. For each GO term, proteins are sampled based on their unique $\mathrm{IC}_{\text{protein function}}$ values (rounded to 3 decimal places). To ensure balanced species representation, a species quota strategy is applied based on the proportions of Eukaryota, Bacteria, Archaea, and Viruses in the dataset after sequence redundancy removal. This ensures an unbiased species distribution in the final sample.
The bilingual deduplication process reduces redundancy in two aspects, amino acid sequence and annotation semantics, ensuring a balanced and diverse protein corpus.

\subsubsection{LLM-based QA Construction}
\label{sec:step3}

To transform curated protein annotations into natural-language question–answer pairs, we prompt the DeepSeek-R1~\cite{guo2025deepseek} model to generate biologically grounded QA texts that reflect both functional attributes and contextual knowledge (the prompts used for each QA type are provided in Appendix~\ref{appendix:llm_prompt}). The resulting QA corpus covers four complementary types:
\blackcircle{1} \textit{Attribute-based QA} captures factual properties directly associated with a protein, such as molecular function, cellular component, or family.
\blackcircle{2} \textit{Knowledge-based QA}  comprises concise, annotation-driven questions and answers that involve in multiple biological aspects of a protein, such as expression, localization, mechanism, and interactions.
\blackcircle{3} \textit{Descriptive Text QA} produces longer natural-language explanations that integrate multiple annotations into coherent functional summaries.
\blackcircle{4} \textit{True or False QA} consists of single statements that integrate multiple biological aspects of a protein, accompanied by a True/False answer and a brief explanation.

These four types yield a rich and varied bilingual dataset, ensuring that models are exposed to both concise factual knowledge and more detailed contextual explanations, supporting their ability to understand and reason about protein functions.

\subsection{Bilingual Contextual Learning}
\label{sec:framwork}
\begin{figure}[ht]
\centering
\includegraphics[width=1\linewidth]{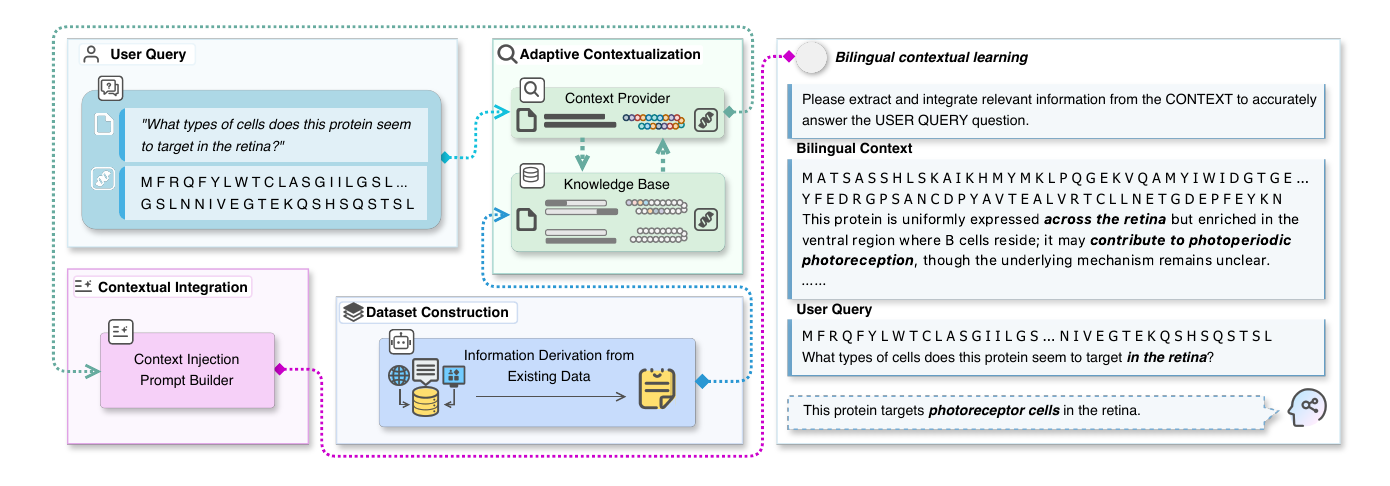}
\caption{\textbf{Process of Query-Adaptive Context Construction. }}
\label{fig:context}
\end{figure}

In practical scenarios, questions concerning protein sequences are often highly flexible and complex: they require not only analogous proteins with similar sequence patterns to capture potential structural or functional signals, but also complementary descriptive knowledge and QA pairs to provide semantic grounding.
As shown in Figure~\ref{fig:context}, we propose an adaptive context construction mechanism, for \textbf{\textit{bilingual contextual learning}}, designed to selectively build bilingual learning contexts for each query. 
Instead of brute-force mixing of amino acid sequences and descriptive texts, the mechanism follows the principle of second language acquisition—exposing learners to new words in context so that meaning and usage can be inferred~\cite{huckin1999incidental}.
By analogy, LLMs acquire protein semantics and reasoning ability through context-driven exposure that grounds sequence patterns in functional and structural exemplars.

The mechanism operates in three stages. First, the adaptive context provider selects candidate contexts from the protein–natural language bilingual dataset, guided by two criteria: 
(i) amino acid sequence homology between candidate proteins and the query sequence, computed with MMseqs2~\cite{steinegger2017mmseqs2}, and (ii) similarity between the descriptive texts or QA pairs of candidate proteins and the query question.
Second, the contextual integration module structures the selected examples into a coherent context. Finally, the constructed bilingual context is combined with the query and presented to the LLM as in-context examples, enabling analogy-based reasoning and evidence integration to produce biologically meaningful responses.
\section{Experiments}

\subsection{setup}
\paragraph{Evaluation Datasets}

We comprehensively evaluated our method using 3 text-based protein understanding datasets: \blackcircle{1} ProtDescribe~\cite{xu2023protst} comprises 553,052 high-quality protein–text pairs extracted from Swiss-Prot. Each instance pairs an amino-acid sequence with a single textual description obtained by concatenating four annotation fields in a fixed order: protein name, function, subcellular location, and similarity. The resulting descriptions average 40–60 tokens.
\blackcircle{2} Protein2Text-QA~\cite{Protein2Text2025} comprises 209,847 open-ended question–answer pairs covering 5,574 unique proteins. Each instance consists of an amino-acid sequence, a free-form question, and a concise answer; all QAs are automatically generated from PubMed abstracts/discussion/introduction sections and presented as conversational natural-language text without fixed templates.
\blackcircle{3} Mol-Instructions~\cite{fang2023mol} comprises 2.04 M instruction instances divided into three major sections: molecule-oriented, protein-oriented, and biomolecular-text. The protein-oriented section alone contributes 505 K instructions covering diverse tasks. Each sample is formatted as a natural-language “instruction–input–output” triplet: the input is a UniProt amino-acid sequence, and the output is a free-text answer tailored to the specific task.

\paragraph{Models} All experiments are conducted under identical prompting protocols. We first evaluate the proposed adaptive context construction method on frozen LLMs, including Qwen2.5-3B~\cite{qwen2.5}, Mistral-7B-Instruct-v0.3~\cite{chaplot2023albert}, Qwen3-14B~\cite{qwen3technicalreport}, Kimi-k2~\cite{team2025kimi}, and GPT-4o~\cite{openai2024gpt4}, to test few-shot and compositional reasoning capabilities, thereby mimicking the dynamics of second language acquisition. In addition, we also evaluate fine-tuned protein-oriented LLMs, such as BioT5-plus-base~\cite{pei2023biot5} and ProLLaMA~\cite{lv2025prollama}, which have been explicitly trained on large-scale protein corpora. These models serve as a baseline for comparison, allowing us to examine the performance gains of our method in general-purpose frozen LLMs relative to specialized protein LLMs.

\paragraph{Metrics} 

We evaluate model outputs using both an automatic metric (ROUGE-L~\cite{lin2002manual}) and human evaluation. ROUGE-L~\cite{lin2002manual}, though widely used for text generation, primarily measures lexical overlap and may not fully capture semantic correctness in protein-related QA. To address this limitation, five evaluators rated the quality of generated answers on a 0–5 scale, where 0 denotes garbled and unreadable content, intermediate scores reflect increasing levels of informativeness and accuracy, and 5 represents fully correct outputs (detailed scoring rubrics are provided in Appendix~\ref{appendix:human_rate}). This combined evaluation provides a more reliable assessment of factual accuracy and overall comprehensibility.

\subsection{Quality of Dataset}
\begin{figure}[ht]
\centering
\includegraphics[width=\linewidth]{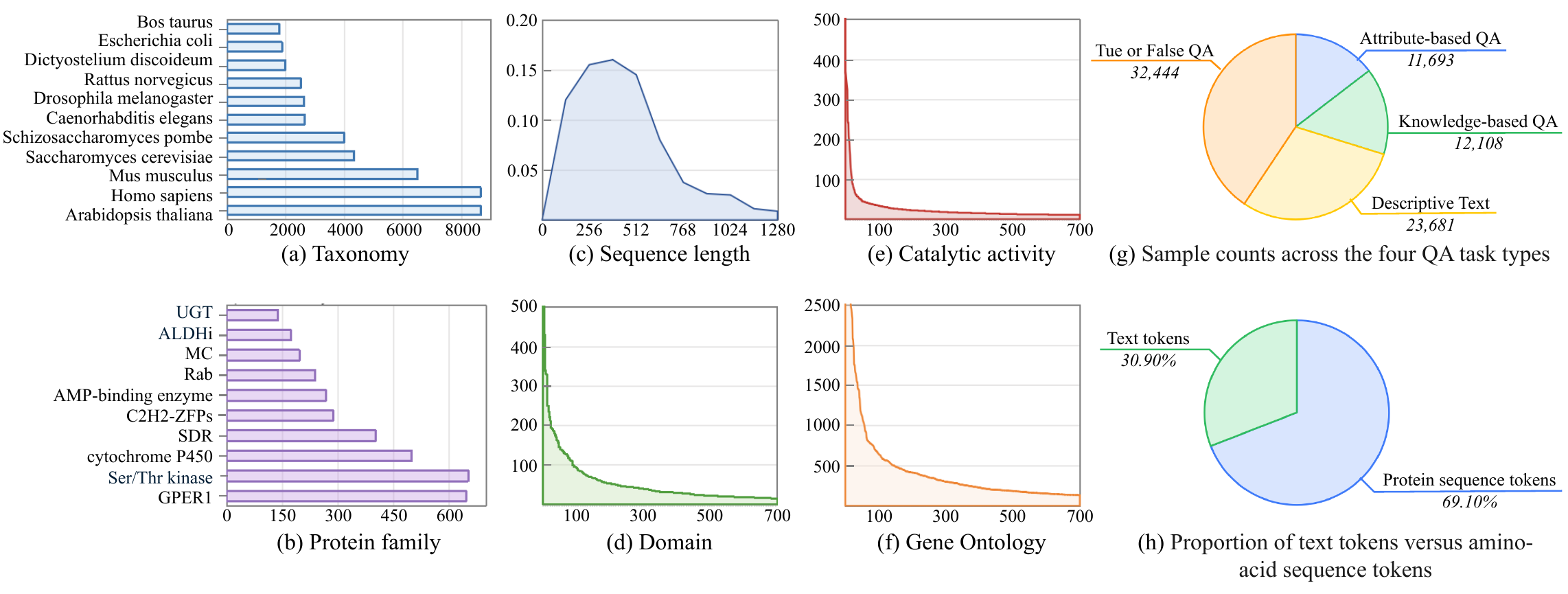}
\caption{\textbf{Dataset statistics.} 
Left: Multidimensional analysis of protein amino-acid sequences, including length, domain composition, and catalytic activity. Right: Sample sizes for the four protein-QA types and the ratio of textual to amino-acid sequence tokens.
}
\label{fig:dataset}
\end{figure}

Figure~\ref{fig:dataset} (a-f) provides a multidimensional analysis of the protein sequences included in our dataset. The collection spans a wide range of sequence lengths, from short peptides to large multi-domain proteins, and covers proteins from 4,135 species across diverse evolutionary lineages. At the family level, the dataset comprises 63,749 families and 1,115 superfamilies, ensuring representation of both well-studied proteins and rare functional groups. Additional annotations capture domain composition, catalytic activity classes, and gene ontology categories, collectively highlighting the long-tail distribution across sequence space and functional categories. 
This diversity ensures broad biological coverage while posing realistic challenges in inferring functions for proteins, particularly for infrequent families and underexplored functions.

Figure~\ref{fig:dataset} (g,h) summarizes the distribution of tasks and token composition within the dataset. The corpus encompasses four distinct protein-QA types, with sample counts ranging from 11,693 (attribute-based QA) to 32,444 (true/false QA), thereby providing balanced coverage across multiple functional perspectives. 
In terms of token composition, amino-acid sequences constitute nearly 70 \% of the corpus, reflecting the sequence-centric nature of protein understanding tasks and highlighting the need for models to align symbolic sequence information with natural-language context effectively.

\subsection{Main Results}
\paragraph{Accuracy gains from context-driven exposure}

Table~\ref{tab:main} presents that our method consistently improves performance on three text-based protein understanding datasets.
Our method raises the average ROUGE-L by 7\% across diverse open-source models and GPT-4o~\cite{openai2024gpt4}, with a maximum gain of 17.2\%, demonstrating that context-driven exposure allows LLMs to acquire protein semantics and reason about function directly from sequence and textual context without any parameter updates. 
Larger models benefit more, suggesting that greater capacity enhances the ability to leverage contextual cues, consistent with learning protein meaning through in-context analogy and reasoning.
In contrast, fine-tuned protein LLMs such as ProLLaMA-7B~\cite{lv2025prollama} do not surpass frozen LLMs augmented with our method, likely due to limited training coverage and task-specific rigidity. This underscores the our method as a lightweight alternative that enables general-purpose LLMs to potentially exceed the performance of domain-adapted models.

\begin{table}[t]
\small
\centering
\setlength{\tabcolsep}{4pt}
\renewcommand\arraystretch{1.2}
    \centering\small
      \caption{\textbf{
  Comparison of different approaches in protein question answering
  } ``Fun.'', ``Des.'', ``Dom.'', and ``Cat.'' denote the 4 protein-oriented tasks in the Mol-Instructions dataset~\cite{fang2023mol}: protein function prediction (Fun.), general textual description generation (Des.), domain/motif recognition (Dom.), and catalytic activity prediction (Cat.). $\Delta$ \textit{Gain} shows the percentage performance increase. $\diamondsuit$ indicates LLMs augmented with our adaptive context construction method. Metric: ROUGE-L.
  }
  \vspace{\baselineskip}
    \begin{tabular}{l|c|c|c@{\hspace{8pt}}c@{\hspace{8pt}}c@{\hspace{8pt}}c@{\hspace{8pt}}c}
\hline

\hline

\hline

\hline
        \multirow{2}{*}{\bf Model}& \multirow{2}{*}{\bf ProtDescribe
} &  \multicolumn{1}{c|}{ \bf Protein2Text-
}&\multicolumn{5}{c}{\bf Mol-Instructions}  \\
          & & \bf QA & Func. & Desc.
 & Dom. & Cat. & \bf Avg.  \\ 
\hline
\rowcolor{gray!20}
        \textit{Fine-tuned LLM} &&&&&&&\\
        BioT5+~\cite{pei2023biot5} & 9.97 & 6.96 & 2.92 & 6.22 & 2.37 & 2.87 & 3.60 \\
        ProLLaMA-7B~\cite{lv2025prollama}& 12.77 & 10.09 & 16.89 & 15.34 & 15.85 & 19.32 & 16.85 \\
\hline
\rowcolor{gray!20} 
        \textit{Frozen LLM} &&&&&&&\\
        Qwen2.5-3B~\cite{qwen2.5} & 18.45 & 23.21 & 18.91 & 17.18 & 18.01 & 20.05 & 18.54 \\
        \rowcolor[HTML]{E6F7FF}
        Qwen2.5-3B~\cite{qwen2.5} $\diamondsuit$& 27.32 & 28.66 & 22.05 & 22.23 & 25.14 & 15.96 & 21.35\\
        $\Delta$ \it Gain & \bf {\color[HTML]{006400} +8.87} & \bf {\color[HTML]{006400} +5.45} & & & & & \bf {\color[HTML]{006400} +2.81}\\
        Mistral-7B-Instruct-v0.3~\cite{chaplot2023albert} & 15.02 & 20.97 & 17.05 & 18.59 & 14.95 & 18.07 & 17.17\\
        \rowcolor[HTML]{E6F7FF}
        Mistral-7B-Instruct-v0.3~\cite{chaplot2023albert} $\diamondsuit$ & 29.39 & 28.59 & 15.77 &22.72 & 17.46 & 21.20 & 19.29\\
        $\Delta$ \it Gain & \bf {\color[HTML]{006400} +14.37} & \bf {\color[HTML]{006400} +7.62} & & & & & \bf {\color[HTML]{006400} +2.12}\\
        Qwen3-14B~\cite{qwen3technicalreport} &  23.20 & 21.02 & 15.80  &  12.75 & 15.81 & 14.06 & 14.61 \\
        \rowcolor[HTML]{E6F7FF}
        Qwen3-14B~\cite{qwen3technicalreport} $\diamondsuit$ &  35.53 & 25.93 & 20.17 & 17.37 & 18.47 & 23.25 & 19.82\\
        $\Delta$ \it Gain & \bf {\color[HTML]{006400}+12.33} & \bf {\color[HTML]{006400} +4.91 } & & & & & \bf {\color[HTML]{006400} +5.21}\\
        kimi-k2~\cite{team2025kimi} & 26.74 & 17.33 & 12.60 &12.36 & 10.32 & 15.97 & 12.81 \\
        \rowcolor[HTML]{E6F7FF}
        kimi-k2~\cite{team2025kimi} $\diamondsuit$ & 35.91 & 21.04 & 14.47 & 14.97 & 15.68 & 17.02 & 15.54\\
        $\Delta$ \it Gain & \bf {\color[HTML]{006400} +9.17} & \bf {\color[HTML]{006400} +3.71} & & & & & \bf {\color[HTML]{006400} +2.72}\\
        GPT-4o~\cite{openai2024gpt4} & 18.29 &20.84 & 16.89 & 14.50 & 16.74 & 20.00 & 17.03\\
        \rowcolor[HTML]{E6F7FF}
        GPT-4o~\cite{openai2024gpt4} $\diamondsuit$ &35.53 & 26.86 & 20.24 & 19.23 & 17.46 & 22.61 & 19.89\\
        $\Delta$ \it Gain & \bf {\color[HTML]{006400} +17.22} & \bf {\color[HTML]{006400} +6.02} & & & & & \bf {\color[HTML]{006400} +2.85}\\
        \cline{1-4}
\hline

\hline

\hline

\hline
    \end{tabular}
    
    \label{tab:main}
\end{table}

\begin{figure}[ht]
\centering
\includegraphics[width=0.9\linewidth]{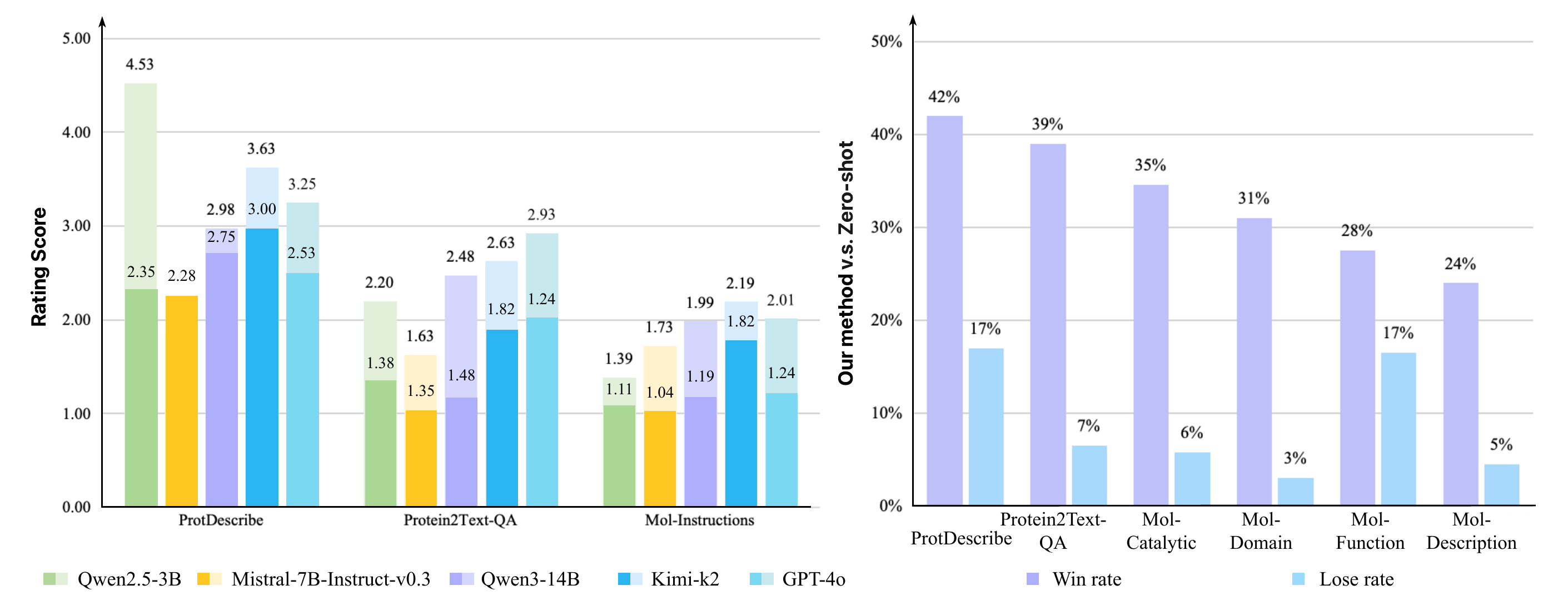}
\caption{\textbf{Comparison of human evaluation results. }
Left: Absolute human rating scores (0–5) for zero-shot model outputs (dark bars) and model outputs with adaptive context exposure (light bars) on three datasets.
Right: Pairwise win/lose proportions comparing outputs with and without adaptive context exposure. Each comparison is based on 8 randomly selected cases per subset (48 cases in total across six subsets).
}
\label{fig:human_rating}
\end{figure}

Human evaluation further demonstrates that exposing models to curated protein–language contexts improves the perceived quality of outputs (Figure~\ref{fig:human_rating}).
Across all rated instances, inter-rater consistency was substantial (Krippendorff’s $\alpha$ = 0.72\%), ensuring reliable annotations; the detailed rubric is given in Appendix~\ref{appendix:human_rate}.
Models receiving context-driven exposures achieve higher or comparable ratings on most tasks (left panel), with the clearest improvements observed on Protein2Text-QA~\cite{Protein2Text2025} and several Mol-Instructions~\cite{fang2023mol} subtasks.
Furthermore, pairwise win/lose analyses (right panel) show that outputs generated with context-driven exposure are preferred in the majority of comparisons, with win rates systematically exceeding loss rates.

\begin{figure}[ht]
\centering
\includegraphics[width=0.72\linewidth]{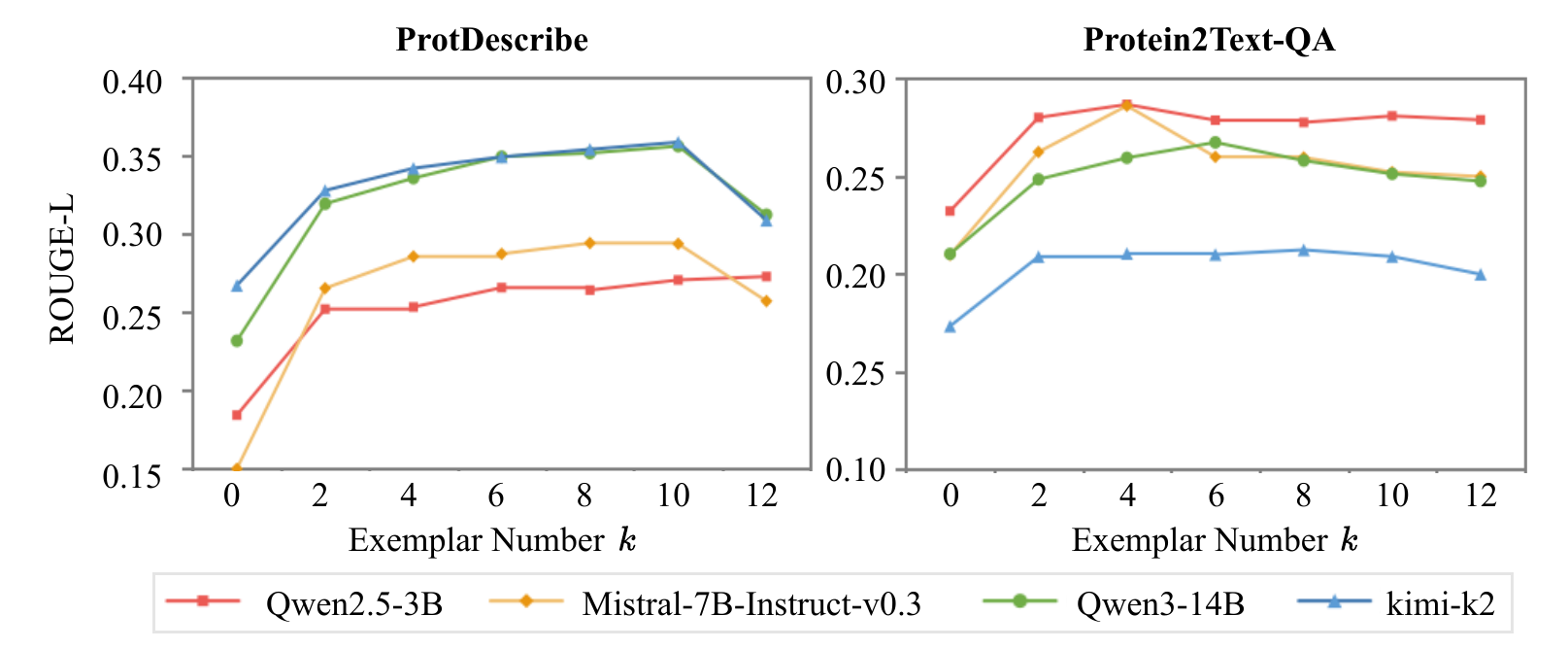}
\caption{\textbf{Effect of varying exemplar number ($k$) on model performance.} We explored $k\in[1,12]$ as the search space; the upper bound was set after a coarse scan up to $k=50$ showed performance saturation around 2-12 exemplars. 
Metric: ROUGE-L.
}
\label{fig:varying_k}
\end{figure}

\paragraph{Varying exemplar number ($k$)}
Figure~\ref{fig:varying_k} illustrates how model performance varies with the number of exemplars ($k$) provided in context across different datasets. Performance generally improves as $k$ increases, but only up to a task-dependent optimum; beyond this point, additional exemplars offer little benefit or even introduce noise. 
The optimal $k$ differs by task. For ProtDescribe~\cite{xu2023protst}, which involves fixed attribute-centric questions, a larger set of bilingual exemplars from related proteins helps the model capture recurring patterns, with performance peaking at $k=10$–$11$. In contrast, Protein2Text-QA~\cite{Protein2Text2025} requires open-ended and integrative reasoning, where only a small number of highly relevant exemplars are beneficial; here, performance peaks earlier at $k=3$–$4$. 
In our experiments, we therefore adopt the task-specific optimal settings: $k=11$ for ProtDescribe~\cite{xu2023protst}, $k=4$ for Protein2Text-QA~\cite{Protein2Text2025}, and $k=4$ for Mol-Instructions~\cite{fang2023mol}. 

\paragraph{Ablation on dual-criterion context selection}
Table~\ref{tab:ablation} shows that using both sequence homology and text/QA similarity (Dual) outperforms either criterion alone, providing complementary signals that maximize the effectiveness of context-driven exemplar selection. On average across three datasets, using only sequence homology reduces performance by 5.2\%, and using only text/QA similarity reduces performance by 2.8\% compared to Dual, though all variants still outperform zero-shot models.

\begin{table}[t]
\small
\centering
\setlength{\tabcolsep}{2pt}
\renewcommand\arraystretch{1.2}
    \centering\small
    \caption{\textbf{Ablation on dual-criterion context selection. } Columns show model performance when using both sequence homology and text/QA similarity (Dual), sequence homology only (SeqOnly), or text/QA similarity only (QAOnly) to guide exemplar selection. Metric: ROUGE-L.
  }
  \vspace{\baselineskip}
    \begin{tabular}{l|ccc|ccc|ccc}
\hline

\hline

\hline

\hline
        \multirow{2}{*}{\bf Model}& \multicolumn{3}{c|}{\bf ProtDescribe} & \multicolumn{3}{c|}{ \bf Protein2Text-QA}&\multicolumn{3}{c}{\bf Mol-Instructions}  \\
         & Dual & SeqOnly & QAOnly & Dual & SeqOnly & QAOnly & Dual & SeqOnly & QAOnly\\ 
\hline
        Qwen2.5-3B~\cite{qwen2.5} & 27.32 & 20.10 & 25.24 & 28.66 & 27.76 & 25.73 & 21.35 & 20.16 & 19.68 \\
        Mistral-7B-Instruct-v0.3~\cite{chaplot2023albert} & 29.39 & 19.46 & 25.24 & 28.59 & 21.85 & 22.68 & 19.12 & 14.31 & 17.40 \\
         Qwen3-14B~\cite{qwen3technicalreport} & 35.53 & 22.90 & 30.52 & 25.93 & 23.26 & 25.87 & 19.82 & 15.07 & 17.58\\
         kimi-k2~\cite{team2025kimi} & 35.91 & 28.58 & 32.43 & 21.04 & 17.10 & 18.06 & 15.54 & 12.79 & 12.09\\
         GPT-4o~\cite{openai2024gpt4} & 35.51 & 27.72 & 32.59 & 26.86 & 23.90 & 26.45 & 19.89 & 17.46 & 17.08\\
        
        \cline{1-4}
\hline

\hline

\hline

\hline
    \end{tabular}
    \label{tab:ablation}
\end{table}

\paragraph{Case studies and qualitative evaluation}

\begin{figure}[ht]
\centering
\includegraphics[width=1\linewidth]{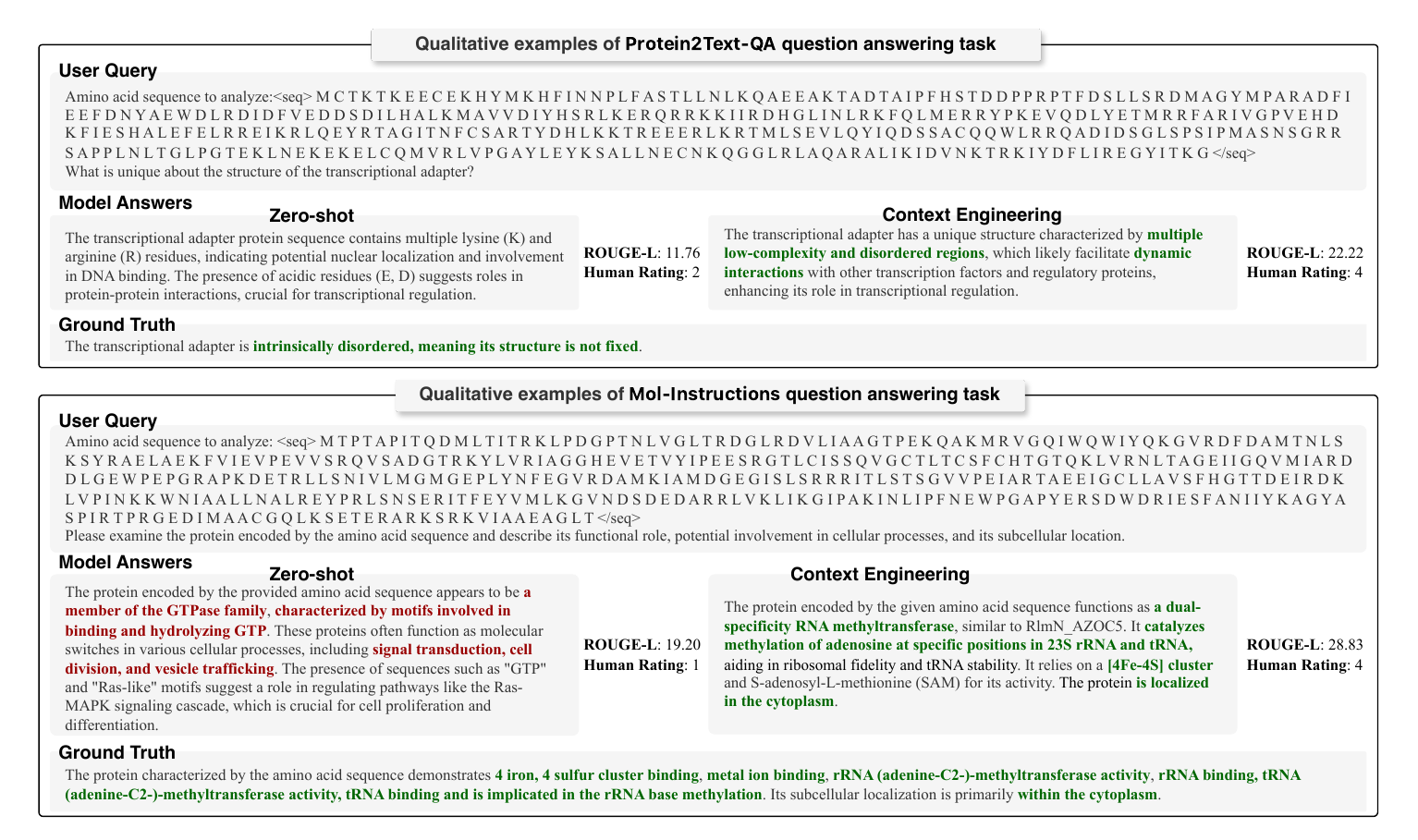}
\caption{\textbf{Qualitative examples of protein question answering.} We present two examples with answers generated by GPT-4o~\cite{openai2024gpt4} along with the target ground truth. The green color highlights accurate keywords, while the red color indicates prediction errors.}
\label{fig:qualitative_examples}
\end{figure}

Figure~\ref{fig:qualitative_examples} illustrates that context-driven exposure produces concise, function-specific descriptions consistent with UniProt annotations. In the two examples shown, the model correctly identifies ``intrinsically disordered regions'', and ``[4Fe-4S] RNA methyltransferase activity'', whereas zero-shot outputs remain generic. 

\section{Conclusion}

We have proposed the “\textbf{\textit{Protein-as-Second-Language}}” framework, which leverages adaptive context construction to enhance bilingual protein understanding by dynamically integrating sequence homology and textual similarity. Additionally, we introduced \textbf{\textit{a protein-natural language bilingual dataset}}, specifically designed to support this framework and facilitate the bridging of protein sequences with functional descriptions. Our approach has successfully enhanced large language models' ability to acquire protein semantics and reasoning capabilities without the need for task-specific parameter updates. Experiments on multiple protein-language datasets demonstrate that our framework consistently outperforms zero-shot baselines, highlighting the effectiveness of context-driven learning in bridging protein sequences with functional descriptions.

\section{Ethics statement}

This work complies with ethical standards and established research practices. All protein data were sourced from publicly available databases, with no proprietary or confidential information involved. Quality assurance and safety checks were applied to minimize harmful or inappropriate content. We acknowledge the broader risks of combining LLMs with biomolecular knowledge, including potential misuse for harmful purposes, and therefore emphasize responsible use guided by fairness, transparency, and accountability. Any harmful or unsafe applications of this dataset are strictly prohibited.

\section{Reproducibility statement}
 
We provide detailed descriptions of the protein–natural language bilingual dataset (Sec.~\ref{sec:dataset}, Appendix~\ref{appendix:dataset_detail}), the adaptive context construction mechanism (Sec.~\ref{sec:framwork}). Data processing steps and QA generation prompts for all four question types are included in Sec.~\ref{sec:dataset} and Appendix~\ref{appendix:llm_prompt}. Code implementing the framework and instructions for reproducing experiments on both frozen and protein-adapted LLMs will be provided as supplementary material upon acceptance.


\begin{thebibliography}{60}
\providecommand{\natexlab}[1]{#1}
\providecommand{\url}[1]{\texttt{#1}}
\expandafter\ifx\csname urlstyle\endcsname\relax
  \providecommand{\doi}[1]{doi: #1}\else
  \providecommand{\doi}{doi: \begingroup \urlstyle{rm}\Url}\fi

\bibitem[Abdine et~al.(2024)Abdine, Chatzianastasis, Bouyioukos, and Vazirgiannis]{abdine2024prot2text}
Hadi Abdine, Michail Chatzianastasis, Costas Bouyioukos, and Michalis Vazirgiannis.
\newblock Prot2text: Multimodal protein's function generation with gnns and transformers.
\newblock In \emph{Proceedings of the AAAI Conference on Artificial Intelligence}, volume~38, pp.\  10757--10765, 2024.

\bibitem[Bairoch \& Apweiler(2000)Bairoch and Apweiler]{bairoch2000swiss}
Amos Bairoch and Rolf Apweiler.
\newblock The swiss-prot protein sequence database and its supplement trembl in 2000.
\newblock \emph{Nucleic acids research}, 28\penalty0 (1):\penalty0 45--48, 2000.

\bibitem[Brandes et~al.(2022)Brandes, Ofer, Peleg, Rappoport, and Linial]{brandes2022proteinbert}
Nadav Brandes, Dan Ofer, Yam Peleg, Nadav Rappoport, and Michal Linial.
\newblock Proteinbert: a universal deep-learning model of protein sequence and function.
\newblock \emph{Bioinformatics}, 38\penalty0 (8):\penalty0 2102--2110, 2022.

\bibitem[Cao \& Shen(2021)Cao and Shen]{cao2021tale}
Yue Cao and Yang Shen.
\newblock Tale: Transformer-based protein function annotation with joint sequence--label embedding.
\newblock \emph{Bioinformatics}, 37\penalty0 (18):\penalty0 2825--2833, 2021.

\bibitem[Chaplot(2023)]{chaplot2023albert}
Devendra~Singh Chaplot.
\newblock Albert q. jiang, alexandre sablayrolles, arthur mensch, chris bamford, devendra singh chaplot, diego de las casas, florian bressand, gianna lengyel, guillaume lample, lucile saulnier, l{\'e}lio renard lavaud, marie-anne lachaux, pierre stock, teven le scao, thibaut lavril, thomas wang, timoth{\'e}e lacroix, william el sayed.
\newblock \emph{arXiv preprint arXiv:2310.06825}, 3, 2023.

\bibitem[Chen et~al.(2024{\natexlab{a}})Chen, Cheng, Li, Geng, Gong, Li, Bei, Tan, Wang, Zeng, et~al.]{chen2024xtrimopglm}
Bo~Chen, Xingyi Cheng, Pan Li, Yangli-ao Geng, Jing Gong, Shen Li, Zhilei Bei, Xu~Tan, Boyan Wang, Xin Zeng, et~al.
\newblock xtrimopglm: unified 100b-scale pre-trained transformer for deciphering the language of protein.
\newblock \emph{arXiv preprint arXiv:2401.06199}, 2024{\natexlab{a}}.

\bibitem[Chen et~al.(2024{\natexlab{b}})Chen, Chen, Xie, Xue, Zhang, Zhou, and Fang]{chen2024unifying}
Zhiyuan Chen, Tianhao Chen, Chenggang Xie, Yang Xue, Xiaonan Zhang, Jingbo Zhou, and Xiaomin Fang.
\newblock Unifying sequences, structures, and descriptions for any-to-any protein generation with the large multimodal model helixprotx.
\newblock \emph{arXiv preprint arXiv:2407.09274}, 2024{\natexlab{b}}.

\bibitem[Clark \& Radivojac(2011)Clark and Radivojac]{clark2011analysis}
Wyatt~T Clark and Predrag Radivojac.
\newblock Analysis of protein function and its prediction from amino acid sequence.
\newblock \emph{Proteins: Structure, Function, and Bioinformatics}, 79\penalty0 (7):\penalty0 2086--2096, 2011.

\bibitem[Devos \& Valencia(2000)Devos and Valencia]{devos2000practical}
Damien Devos and Alfonso Valencia.
\newblock Practical limits of function prediction.
\newblock \emph{Proteins: Structure, Function, and Bioinformatics}, 41\penalty0 (1):\penalty0 98--107, 2000.

\bibitem[Elnaggar et~al.(2021)Elnaggar, Heinzinger, Dallago, Rehawi, Wang, Jones, Gibbs, Feher, Angerer, Steinegger, et~al.]{elnaggar2021prottrans}
Ahmed Elnaggar, Michael Heinzinger, Christian Dallago, Ghalia Rehawi, Yu~Wang, Llion Jones, Tom Gibbs, Tamas Feher, Christoph Angerer, Martin Steinegger, et~al.
\newblock Prottrans: Toward understanding the language of life through self-supervised learning.
\newblock \emph{IEEE transactions on pattern analysis and machine intelligence}, 44\penalty0 (10):\penalty0 7112--7127, 2021.

\bibitem[Fang et~al.(2024)Fang, Liang, Zhang, Liu, Huang, Chen, Fan, and Chen]{fang2023mol}
Yin Fang, Xiaozhuan Liang, Ningyu Zhang, Kangwei Liu, Rui Huang, Zhuo Chen, Xiaohui Fan, and Huajun Chen.
\newblock Mol-instructions: {A} large-scale biomolecular instruction dataset for large language models.
\newblock In \emph{The Twelfth International Conference on Learning Representations}. OpenReview.net, 2024.
\newblock URL \url{https://openreview.net/pdf?id=Tlsdsb6l9n}.

\bibitem[Ferruz et~al.(2022)Ferruz, Schmidt, and H{\"o}cker]{ferruz2022protgpt2}
Noelia Ferruz, Steffen Schmidt, and Birte H{\"o}cker.
\newblock Protgpt2 is a deep unsupervised language model for protein design.
\newblock \emph{Nature communications}, 13\penalty0 (1):\penalty0 4348, 2022.

\bibitem[Gass et~al.(2020)Gass, Behney, and Plonsky]{gass2020second}
Susan~M Gass, Jennifer Behney, and Luke Plonsky.
\newblock \emph{Second language acquisition: An introductory course}.
\newblock Routledge, 2020.

\bibitem[Guo et~al.(2025)Guo, Yang, Zhang, Song, Zhang, Xu, Zhu, Ma, Wang, Bi, et~al.]{guo2025deepseek}
Daya Guo, Dejian Yang, Haowei Zhang, Junxiao Song, Ruoyu Zhang, Runxin Xu, Qihao Zhu, Shirong Ma, Peiyi Wang, Xiao Bi, et~al.
\newblock Deepseek-r1: Incentivizing reasoning capability in llms via reinforcement learning.
\newblock \emph{arXiv preprint arXiv:2501.12948}, 2025.

\bibitem[Guo et~al.(2023)Guo, Huo, Zhang, and Xie]{guo2023proteinchat}
Han Guo, Mingjia Huo, Ruiyi Zhang, and Pengtao Xie.
\newblock Proteinchat: Towards achieving chatgpt-like functionalities on protein 3d structures.
\newblock \emph{Authorea Preprints}, 2023.

\bibitem[Hayes et~al.(2025)Hayes, Rao, Akin, Sofroniew, Oktay, Lin, Verkuil, Tran, Deaton, Wiggert, et~al.]{hayes2025simulating}
Thomas Hayes, Roshan Rao, Halil Akin, Nicholas~J Sofroniew, Deniz Oktay, Zeming Lin, Robert Verkuil, Vincent~Q Tran, Jonathan Deaton, Marius Wiggert, et~al.
\newblock Simulating 500 million years of evolution with a language model.
\newblock \emph{Science}, pp.\  eads0018, 2025.

\bibitem[Hu et~al.(2024)Hu, Tan, Xu, Gao, Xia, Wu, and Li]{ProtGO}
Bozhen Hu, Cheng Tan, Yongjie Xu, Zhangyang Gao, Jun Xia, Lirong Wu, and Stan~Z. Li.
\newblock Protgo: Function-guided protein modeling for unified representation learning.
\newblock In A.~Globerson, L.~Mackey, D.~Belgrave, A.~Fan, U.~Paquet, J.~Tomczak, and C.~Zhang (eds.), \emph{Advances in Neural Information Processing Systems}, volume~37, pp.\  88581--88604. Curran Associates, Inc., 2024.
\newblock URL \url{https://proceedings.neurips.cc/paper_files/paper/2024/file/a1722a6bd1023c026a3d6a570fb3af75-Paper-Conference.pdf}.

\bibitem[Huckin \& Coady(1999)Huckin and Coady]{huckin1999incidental}
Thomas Huckin and James Coady.
\newblock Incidental vocabulary acquisition in a second language: A review.
\newblock \emph{Studies in second language acquisition}, 21\penalty0 (2):\penalty0 181--193, 1999.

\bibitem[Jararweh et~al.(2025)Jararweh, Macaulay, Arredondo, Hu, Tafoya, Virupakshappa, and Sahu]{Protein2Text2025}
Ala Jararweh, Oladimeji Macaulay, David Arredondo, Yue Hu, Luis Tafoya, Kushal Virupakshappa, and Avinash Sahu.
\newblock Protein2text: Resampling mechanism to translate protein sequences into human-interpretable text.
\newblock In \emph{NAACL 2025 - Industry Track}, 2025.

\bibitem[Jarvis \& Pavlenko(2008)Jarvis and Pavlenko]{jarvis2008crosslinguistic}
Scott Jarvis and Aneta Pavlenko.
\newblock \emph{Crosslinguistic influence in language and cognition}.
\newblock Routledge, 2008.

\bibitem[Kirkpatrick et~al.(2017)Kirkpatrick, Pascanu, Rabinowitz, Veness, Desjardins, Rusu, Milan, Quan, Ramalho, Grabska-Barwinska, et~al.]{kirkpatrick2017overcoming}
James Kirkpatrick, Razvan Pascanu, Neil Rabinowitz, Joel Veness, Guillaume Desjardins, Andrei~A Rusu, Kieran Milan, John Quan, Tiago Ramalho, Agnieszka Grabska-Barwinska, et~al.
\newblock Overcoming catastrophic forgetting in neural networks.
\newblock \emph{Proceedings of the national academy of sciences}, 114\penalty0 (13):\penalty0 3521--3526, 2017.

\bibitem[Kitadai \& Maruyama(2018)Kitadai and Maruyama]{kitadai2018origins}
Norio Kitadai and Shigenori Maruyama.
\newblock Origins of building blocks of life: A review.
\newblock \emph{Geoscience Frontiers}, 9\penalty0 (4):\penalty0 1117--1153, 2018.

\bibitem[Koonin \& Galperin(2002)Koonin and Galperin]{koonin2002sequence}
Eugene Koonin and Michael~Y Galperin.
\newblock \emph{Sequence—Evolution—Function: Computational Approaches in Comparative Genomics}.
\newblock Springer Science \& Business Media, 2002.

\bibitem[Lin \& Hovy(2002)Lin and Hovy]{lin2002manual}
Chin-Yew Lin and Eduard Hovy.
\newblock Manual and automatic evaluation of summaries.
\newblock In \emph{Proceedings of the ACL-02 workshop on automatic summarization}, pp.\  45--51, 2002.

\bibitem[Lin et~al.(2023)Lin, Akin, Rao, Hie, Zhu, Lu, Smetanin, Verkuil, Kabeli, Shmueli, et~al.]{lin2023evolutionary}
Zeming Lin, Halil Akin, Roshan Rao, Brian Hie, Zhongkai Zhu, Wenting Lu, Nikita Smetanin, Robert Verkuil, Ori Kabeli, Yaniv Shmueli, et~al.
\newblock Evolutionary-scale prediction of atomic-level protein structure with a language model.
\newblock \emph{Science}, 379\penalty0 (6637):\penalty0 1123--1130, 2023.

\bibitem[Liu et~al.(2024{\natexlab{a}})Liu, Sun, Ji, Tian, Tang, Wu, and Lan]{liu2024evollama}
Nuowei Liu, Changzhi Sun, Tao Ji, Junfeng Tian, Jianxin Tang, Yuanbin Wu, and Man Lan.
\newblock Evollama: Enhancing llms' understanding of proteins via multimodal structure and sequence representations.
\newblock \emph{arXiv preprint arXiv:2412.11618}, 2024{\natexlab{a}}.

\bibitem[Liu et~al.(2024{\natexlab{b}})Liu, Zhang, Fei, Zhang, Wang, Kawaguchi, and Chua]{liu-etal-2024-prott3}
Zhiyuan Liu, An~Zhang, Hao Fei, Enzhi Zhang, Xiang Wang, Kenji Kawaguchi, and Tat-Seng Chua.
\newblock {P}rot{T}3: Protein-to-text generation for text-based protein understanding.
\newblock In Lun-Wei Ku, Andre Martins, and Vivek Srikumar (eds.), \emph{Proceedings of the 62nd Annual Meeting of the Association for Computational Linguistics (Volume 1: Long Papers)}, pp.\  5949--5966, Bangkok, Thailand, 2024{\natexlab{b}}. Association for Computational Linguistics.
\newblock \doi{10.18653/v1/2024.acl-long.324}.
\newblock URL \url{https://aclanthology.org/2024.acl-long.324/}.

\bibitem[Luo et~al.(2024)Luo, Zhang, Fan, Yang, Hong, Wu, Qiao, and Nie]{luo2024biomedgpt}
Yizhen Luo, Jiahuan Zhang, Siqi Fan, Kai Yang, Massimo Hong, Yushuai Wu, Mu~Qiao, and Zaiqing Nie.
\newblock Biomedgpt: An open multimodal large language model for biomedicine.
\newblock \emph{IEEE Journal of Biomedical and Health Informatics}, 2024.

\bibitem[Lv et~al.(2024)Lv, Lin, Li, Liu, Cui, Yu-Chian~Chen, Yuan, and Tian]{lv2024prollama}
Liuzhenghao Lv, Zongying Lin, Hao Li, Yuyang Liu, Jiaxi Cui, Calvin Yu-Chian~Chen, Li~Yuan, and Yonghong Tian.
\newblock Prollama: A protein large language model for multi-task protein language processing.
\newblock \emph{arXiv e-prints}, pp.\  arXiv--2402, 2024.

\bibitem[Lv et~al.(2025)Lv, Lin, Li, Liu, Cui, Chen, Yuan, and Tian]{lv2025prollama}
Liuzhenghao Lv, Zongying Lin, Hao Li, Yuyang Liu, Jiaxi Cui, Calvin Yu-Chian Chen, Li~Yuan, and Yonghong Tian.
\newblock Prollama: A protein large language model for multi-task protein language processing.
\newblock \emph{IEEE Transactions on Artificial Intelligence}, 2025.

\bibitem[Ma et~al.(2025)Ma, Fan, Wang, Chen, Lin, Li, Feng, Zhang, Cao, and Gao]{ma2025prottex}
Zicheng Ma, Chuanliu Fan, Zhicong Wang, Zhenyu Chen, Xiaohan Lin, Yanheng Li, Shihao Feng, Jun Zhang, Ziqiang Cao, and Yi~Qin Gao.
\newblock Prottex: Structure-in-context reasoning and editing of proteins with large language models.
\newblock \emph{arXiv preprint arXiv:2503.08179}, 2025.

\bibitem[Madani et~al.(2023)Madani, Krause, Greene, Subramanian, Mohr, Holton, Olmos, Xiong, Sun, Socher, et~al.]{madani2023large}
Ali Madani, Ben Krause, Eric~R Greene, Subu Subramanian, Benjamin~P Mohr, James~M Holton, Jose~Luis Olmos, Caiming Xiong, Zachary~Z Sun, Richard Socher, et~al.
\newblock Large language models generate functional protein sequences across diverse families.
\newblock \emph{Nature echnology}, 41\penalty0 (8):\penalty0 1099--1106, 2023.

\bibitem[Mondon(1984)]{mondon1984classification}
Camille Mondon.
\newblock Classification and regression trees, 1984.

\bibitem[Nijkamp et~al.(2023)Nijkamp, Ruffolo, Weinstein, Naik, and Madani]{nijkamp2023progen2}
Erik Nijkamp, Jeffrey~A Ruffolo, Eli~N Weinstein, Nikhil Naik, and Ali Madani.
\newblock Progen2: exploring the boundaries of protein language models.
\newblock \emph{Cell systems}, 14\penalty0 (11):\penalty0 968--978, 2023.

\bibitem[OpenAI et~al.(2024)]{openai2024gpt4}
OpenAI et~al.
\newblock Gpt-4 technical report, 2024.

\bibitem[Pei et~al.(2023{\natexlab{a}})Pei, Zhang, Zhu, Wu, Gao, Wu, Xia, and Yan]{BioT5}
Qizhi Pei, Wei Zhang, Jinhua Zhu, Kehan Wu, Kaiyuan Gao, Lijun Wu, Yingce Xia, and Rui Yan.
\newblock {B}io{T}5: Enriching cross-modal integration in biology with chemical knowledge and natural language associations.
\newblock In Houda Bouamor, Juan Pino, and Kalika Bali (eds.), \emph{Proceedings of the 2023 Conference on Empirical Methods in Natural Language Processing}, pp.\  1102--1123, Singapore, December 2023{\natexlab{a}}. Association for Computational Linguistics.
\newblock \doi{10.18653/v1/2023.emnlp-main.70}.
\newblock URL \url{https://aclanthology.org/2023.emnlp-main.70/}.

\bibitem[Pei et~al.(2023{\natexlab{b}})Pei, Zhang, Zhu, Wu, Gao, Wu, Xia, and Yan]{pei2023biot5}
Qizhi Pei, Wei Zhang, Jinhua Zhu, Kehan Wu, Kaiyuan Gao, Lijun Wu, Yingce Xia, and Rui Yan.
\newblock Biot5: Enriching cross-modal integration in biology with chemical knowledge and natural language associations.
\newblock \emph{arXiv preprint arXiv:2310.07276}, 2023{\natexlab{b}}.

\bibitem[Pesquita et~al.(2008)Pesquita, Faria, Bastos, Ferreira, Falc{\~a}o, and Couto]{pesquita2008metrics}
Catia Pesquita, Daniel Faria, Hugo Bastos, Ant{\'o}nio~EN Ferreira, Andr{\'e}~O Falc{\~a}o, and Francisco~M Couto.
\newblock Metrics for go based protein semantic similarity: a systematic evaluation.
\newblock \emph{BMC bioinformatics}, 9\penalty0 (Suppl 5):\penalty0 S4, 2008.

\bibitem[Rost et~al.(1998)]{rost1998protein}
Burkhard Rost et~al.
\newblock Protein structure prediction in 1d, 2d, and 3d.
\newblock \emph{Encyclopedia of Computational Chemistry}, pp.\  2242--2255, 1998.

\bibitem[Shen et~al.(2024)Shen, Chen, Mamalakis, He, Xia, Li, Su, He, and Wang]{shen2024fine}
Yiqing Shen, Zan Chen, Michail Mamalakis, Luhan He, Haiyang Xia, Tianbin Li, Yanzhou Su, Junjun He, and Yu~Guang Wang.
\newblock A fine-tuning dataset and benchmark for large language models for protein understanding.
\newblock In \emph{2024 IEEE International Conference on Bioinformatics and Biomedicine (BIBM)}, pp.\  2390--2395. IEEE, 2024.

\bibitem[Steinegger \& S{\"o}ding(2017)Steinegger and S{\"o}ding]{steinegger2017mmseqs2}
Martin Steinegger and Johannes S{\"o}ding.
\newblock Mmseqs2 enables sensitive protein sequence searching for the analysis of massive data sets.
\newblock \emph{Nature biotechnology}, 35\penalty0 (11):\penalty0 1026--1028, 2017.

\bibitem[Su et~al.(2023)Su, Han, Zhou, Shan, Zhou, and Yuan]{su2023saprot}
Jin Su, Chenchen Han, Yuyang Zhou, Junjie Shan, Xibin Zhou, and Fajie Yuan.
\newblock Saprot: Protein language modeling with structure-aware vocabulary.
\newblock \emph{BioRxiv}, pp.\  2023--10, 2023.

\bibitem[Tan et~al.(2025)Tan, Gou, Zhong, Hong, Yu, and Zhou]{tan2025venusxunlockingfinegrainedfunctional}
Yang Tan, Wenrui Gou, Bozitao Zhong, Liang Hong, Huiqun Yu, and Bingxin Zhou.
\newblock Venusx: Unlocking fine-grained functional understanding of proteins, 2025.
\newblock URL \url{https://arxiv.org/abs/2505.11812}.

\bibitem[Taylor et~al.(2022)Taylor, Kardas, Cucurull, Scialom, Hartshorn, Saravia, Poulton, Kerkez, and Stojnic]{taylor2022galactica}
Ross Taylor, Marcin Kardas, Guillem Cucurull, Thomas Scialom, Anthony Hartshorn, Elvis Saravia, Andrew Poulton, Viktor Kerkez, and Robert Stojnic.
\newblock Galactica: A large language model for science.
\newblock \emph{ArXiv preprint}, abs/2211.09085, 2022.
\newblock URL \url{https://arxiv.org/abs/2211.09085}.

\bibitem[Team et~al.(2025)Team, Bai, Bao, Chen, Chen, Chen, Chen, Chen, Chen, Chen, et~al.]{team2025kimi}
Kimi Team, Yifan Bai, Yiping Bao, Guanduo Chen, Jiahao Chen, Ningxin Chen, Ruijue Chen, Yanru Chen, Yuankun Chen, Yutian Chen, et~al.
\newblock Kimi k2: Open agentic intelligence.
\newblock \emph{arXiv preprint arXiv:2507.20534}, 2025.

\bibitem[Team(2024)]{qwen2.5}
Qwen Team.
\newblock Qwen2.5: A party of foundation models, September 2024.
\newblock URL \url{https://qwenlm.github.io/blog/qwen2.5/}.

\bibitem[Team(2025)]{qwen3technicalreport}
Qwen Team.
\newblock Qwen3 technical report, 2025.
\newblock URL \url{https://arxiv.org/abs/2505.09388}.

\bibitem[Wang et~al.(2024)Wang, Fan, Quan, and Yang]{wang2024protchatgpt}
Chao Wang, Hehe Fan, Ruijie Quan, and Yi~Yang.
\newblock Protchatgpt: Towards understanding proteins with large language models.
\newblock \emph{arXiv preprint arXiv:2402.09649}, 2024.

\bibitem[Wang \& Pollock(2005)Wang and Pollock]{wang2005context}
Zhengyuan~O Wang and David~D Pollock.
\newblock Context dependence and coevolution among amino acid residues in proteins.
\newblock In \emph{Methods in enzymology}, volume 395, pp.\  779--790. Elsevier, 2005.

\bibitem[Wu et~al.(2025)Wu, Liu, Cao, Li, Feng, Shu, Yu, Yuan, and Li]{wu2025rethinking}
Juntong Wu, Zijing Liu, He~Cao, Hao Li, Bin Feng, Zishan Shu, Ke~Yu, Li~Yuan, and Yu~Li.
\newblock Rethinking text-based protein understanding: Retrieval or llm?
\newblock \emph{arXiv preprint arXiv:2505.20354}, 2025.

\bibitem[Wu et~al.(2024{\natexlab{a}})Wu, Chang, and Zou]{wu2024proteinclip}
Kevin~E Wu, Howard Chang, and James Zou.
\newblock Proteinclip: enhancing protein language models with natural language.
\newblock \emph{bioRxiv}, pp.\  2024--05, 2024{\natexlab{a}}.

\bibitem[Wu et~al.(2024{\natexlab{b}})Wu, Yost, Daniel, Belk, Xia, Egawa, Satpathy, Chang, and Zou]{wu2024tcr}
Kevin~E Wu, Kathryn Yost, Bence Daniel, Julia Belk, Yu~Xia, Takeshi Egawa, Ansuman Satpathy, Howard Chang, and James Zou.
\newblock Tcr-bert: learning the grammar of t-cell receptors for flexible antigen-binding analyses.
\newblock In \emph{Machine Learning in Computational Biology}, pp.\  194--229. PMLR, 2024{\natexlab{b}}.

\bibitem[Xiang et~al.(2024)Xiang, Xiong, Chen, Xiong, Zhang, Fu, Zheng, Liu, and Shi]{xiang2024fapm}
Wenkai Xiang, Zhaoping Xiong, Huan Chen, Jiacheng Xiong, Wei Zhang, Zunyun Fu, Mingyue Zheng, Bing Liu, and Qian Shi.
\newblock Fapm: functional annotation of proteins using multimodal models beyond structural modeling.
\newblock \emph{Bioinformatics}, 40\penalty0 (12):\penalty0 btae680, 2024.

\bibitem[Xiao et~al.(2024)Xiao, Sun, Jin, Wang, and Wang]{xiao2024proteingpt}
Yijia Xiao, Edward Sun, Yiqiao Jin, Qifan Wang, and Wei Wang.
\newblock Proteingpt: Multimodal llm for protein property prediction and structure understanding.
\newblock \emph{arXiv preprint arXiv:2408.11363}, 2024.

\bibitem[Xiao et~al.(2025)Xiao, Zhao, Zhang, Jin, Zhang, Ren, Sun, Wang, Wan, Lu, et~al.]{xiao2025protein}
Yijia Xiao, Wanjia Zhao, Junkai Zhang, Yiqiao Jin, Han Zhang, Zhicheng Ren, Renliang Sun, Haixin Wang, Guancheng Wan, Pan Lu, et~al.
\newblock Protein large language models: A comprehensive survey.
\newblock \emph{arXiv preprint arXiv:2502.17504}, 2025.

\bibitem[Xu \& Wang(2022)Xu and Wang]{xu2022protranslator}
Hanwen Xu and Sheng Wang.
\newblock Protranslator: zero-shot protein function prediction using textual description.
\newblock In \emph{International conference on research in computational molecular biology}, pp.\  279--294. Springer, 2022.

\bibitem[Xu et~al.(2023{\natexlab{a}})Xu, Yuan, Miret, and Tang]{protst}
Minghao Xu, Xinyu Yuan, Santiago Miret, and Jian Tang.
\newblock Protst: Multi-modality learning of protein sequences and biomedical texts.
\newblock In Andreas Krause, Emma Brunskill, Kyunghyun Cho, Barbara Engelhardt, Sivan Sabato, and Jonathan Scarlett (eds.), \emph{International Conference on Machine Learning, {ICML} 2023, 23-29 July 2023, Honolulu, Hawaii, {USA}}, volume 202 of \emph{Proceedings of Machine Learning Research}, pp.\  38749--38767. {PMLR}, 2023{\natexlab{a}}.
\newblock URL \url{https://proceedings.mlr.press/v202/xu23t.html}.

\bibitem[Xu et~al.(2023{\natexlab{b}})Xu, Yuan, Miret, and Tang]{xu2023protst}
Minghao Xu, Xinyu Yuan, Santiago Miret, and Jian Tang.
\newblock Protst: Multi-modality learning of protein sequences and biomedical texts.
\newblock In \emph{International Conference on Machine Learning}, pp.\  38749--38767. PMLR, 2023{\natexlab{b}}.

\bibitem[Zhang et~al.(2022)Zhang, Bi, Liang, Cheng, Hong, Deng, Zhang, Lian, and Chen]{zhang2022ontoprotein}
Ningyu Zhang, Zhen Bi, Xiaozhuan Liang, Siyuan Cheng, Haosen Hong, Shumin Deng, Qiang Zhang, Jiazhang Lian, and Huajun Chen.
\newblock Ontoprotein: Protein pretraining with gene ontology embedding.
\newblock In \emph{The Tenth International Conference on Learning Representations, {ICLR} 2022, Virtual Event, April 25-29, 2022}. OpenReview.net, 2022.
\newblock URL \url{https://openreview.net/forum?id=yfe1VMYAXa4}.

\bibitem[Zhuo et~al.(2024)Zhuo, Chi, Xu, Huang, Zhao, Zheng, He, Mao, and Zhang]{zhuo2024protllm}
Le~Zhuo, Zewen Chi, Minghao Xu, Heyan Huang, Jianan Zhao, Heqi Zheng, Conghui He, Xian-Ling Mao, and Wentao Zhang.
\newblock {P}rot{LLM}: An interleaved protein-language {LLM} with protein-as-word pre-training.
\newblock In Lun-Wei Ku, Andre Martins, and Vivek Srikumar (eds.), \emph{Proceedings of the 62nd Annual Meeting of the Association for Computational Linguistics (Volume 1: Long Papers)}, pp.\  8950--8963, Bangkok, Thailand, August 2024. Association for Computational Linguistics.
\newblock \doi{10.18653/v1/2024.acl-long.484}.
\newblock URL \url{https://aclanthology.org/2024.acl-long.484/}.

\end{thebibliography}
\bibliographystyle{iclr2026_conference}

\appendix

\section{Evaluation Metrics}
\label{appendix:human_rate}
We use the automatic metric ROUGE-L~\cite{lin2002manual} to assess the quality of the generated text by comparing it with reference answers. In addition, we incorporate manual checking into the evaluation pipeline and compute a human-rating score. Five evaluators with biological-research experience were asked to rate each generated answer on a 0–5 scale (the integer score corresponds to the category number minus one). All evaluators have at least two years of research experience in biology. The six ordinal categories they used are:

\begin{enumerate}
  \item \textbf{Garbled} -- the text is incomprehensible and lacks any readability.
  \item \textbf{Inaccurate} -- the text is readable but entirely incorrect and devoid of meaningful information.
  \item \textbf{Partially informative} -- the text offers some reference value, yet its factual correctness is poor.
  \item \textbf{Moderately accurate} -- roughly half of the information is correct, but several errors remain.
  \item \textbf{Mostly accurate} -- the content is almost entirely correct, with only minor omissions or errors.
  \item \textbf{Completely correct} -- the content is accurate in its entirety, without any mistakes.
\end{enumerate}

\section{ADDITIONAL RESULTS}

\begin{figure}[ht]
\centering
\includegraphics[width=\linewidth]{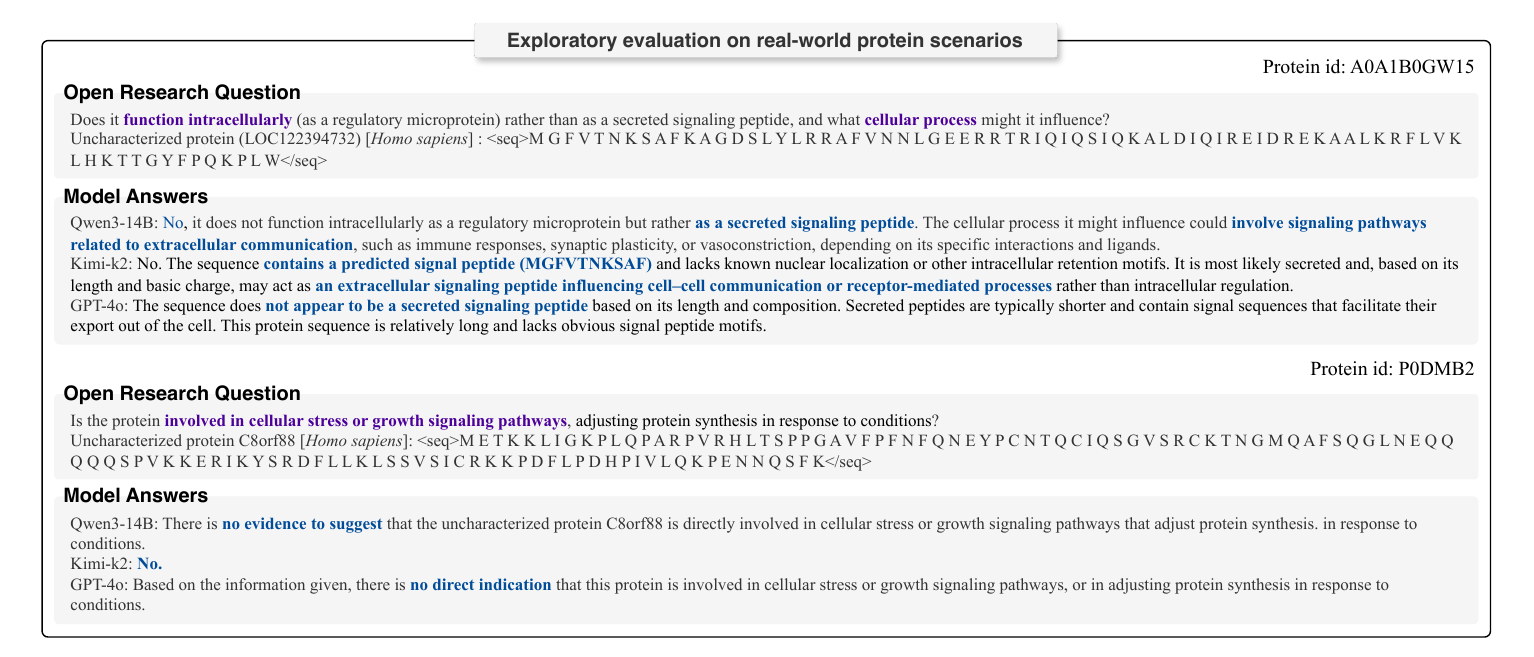}
\caption{\textbf{Exploratory evaluation of our framework on real-world protein scenarios.}  
}
\label{fig:realworld_study}
\end{figure}

\paragraph{Evaluation on real-world protein scenarios} 
To examine the applicability of our framework beyond benchmark datasets, we evaluated it on biologically relevant queries involving uncharacterized \textit{Homo sapiens} proteins. For each case, a current biologically relevant question of research interest was paired with the corresponding protein amino acid sequence and input to representative LLMs guided by our framework. As illustrated in Figure~\ref{fig:realworld_study}, the models produced plausible hypotheses aligned with biological knowledge. These results demonstrate that our framework can extend to real-world scenarios, offering interpretable preliminary insights into proteins lacking experimental annotation and potentially guiding future biological investigations.

\section{Bilingual Dataset: Details and Examples}
\label{appendix:dataset_detail}

\begin{figure}[htbp]
  \centering
  \includegraphics[width=\linewidth]{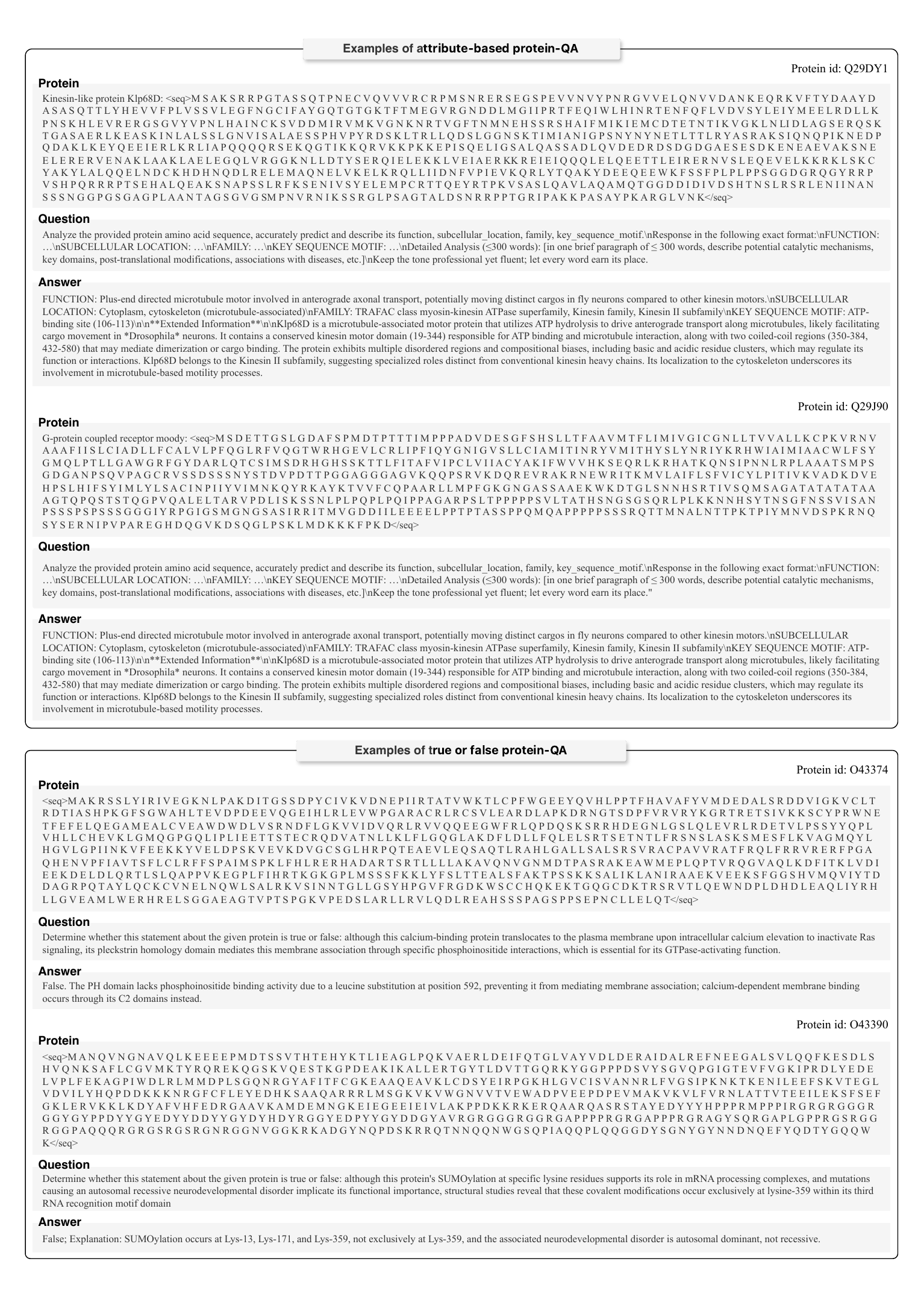}
  \caption{\textbf{Illustrative examples of the four bilingual QA types (Part 1).}}
  \label{fig:dataset_example_p1}
\end{figure}

\begin{figure}[htbp]
  \centering
  \includegraphics[width=\linewidth]{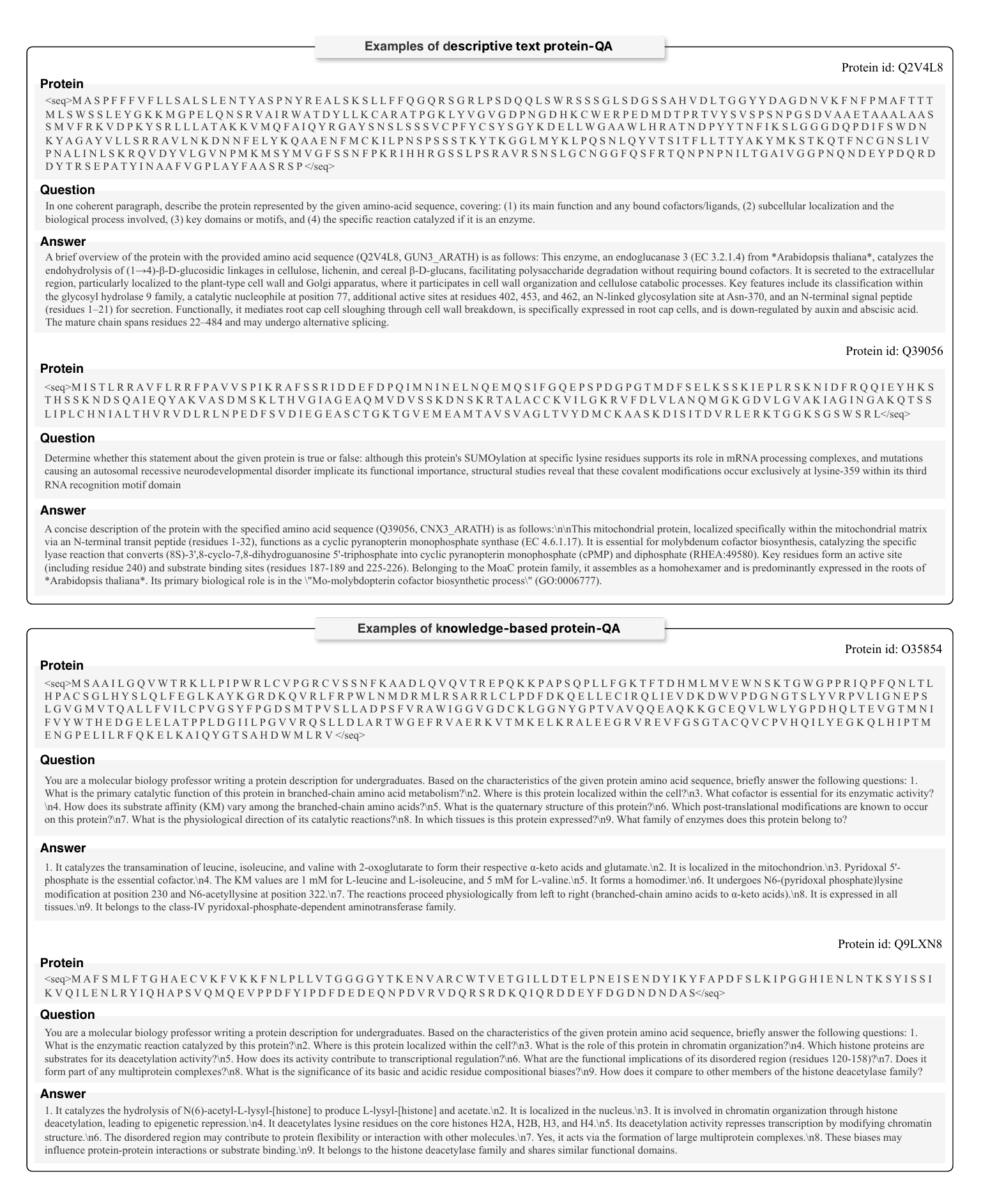}
  \caption{\textbf{Illustrative examples of the four bilingual QA types (Part 2).}}
  \label{fig:dataset_example_p2}
\end{figure}

All protein entries are derived from the manually curated SwissProt~\cite{bairoch2000swiss} section of UniProt, which provides high-quality annotations of protein sequences and functional descriptions. After deduplication, we prompted LLMs to generate four types of bilingual QA pairs from these entries. To assess data quality, we randomly sampled 500 pairs from the full set of 79,926 automatically constructed examples. Each sampled pair was manually reviewed by domain experts along three dimensions: semantic fidelity, biological plausibility, and translation fluency. The review showed a 95\% pass rate, confirming that the dataset maintains high linguistic accuracy and biological reliability. Examples of four bilingual QA types are provided in Figure~\ref{fig:dataset_example_p1} and Figure~\ref{fig:dataset_example_p2}.

\section{llm statement}
\label{appendix:llm_prompt}
We acknowledge the use of LLMs in this work. Specifically, DeepSeek-R1~\cite{guo2025deepseek} was employed for two purposes: (i) polishing the English presentation of the manuscript, and (ii) generating bilingual dataset entries from curated protein annotations, where the prompts were carefully designed to ensure scientific accuracy and linguistic quality. Below we provide the exact prompts used for each bilingual QA type in the dataset construction process. 

Prompt for \texttt{Attribute-based Answer} generation is following:
\begin{scriptsize}\begin{verbatim}
"Based on the provided annotations, compose a concise protein information description in the 
following fixed format: 
PROTEIN NAME: …
FUNCTION: …
SUBCELLULAR LOCATION: …
FAMILY: …
KEY SEQUENCE MOTIF: … (write N/A if none).
After the fixed fields, leave one blank line and proceed to the `Extended Information' 
paragraph. In fluent, professional English, supply any additional details essential for 
understanding the protein, integrating all relevant annotation content in a coherent 
narrative. Maintain brevity and avoid redundancy."
\end{verbatim}\end{scriptsize}

Prompt for \texttt{True or False QA} generation is following:
\begin{scriptsize}\begin{verbatim}
"You are a protein science expert. Please read the UniProt entry above and design 1 True/False 
question that meets all of the following rules: 
(1) The stem must weave together diverse distinct knowledge dimensions from the entry (e.g.,
catalytic chemistry, structural biology, disease relevance, evolutionary conservation, PTM,
mutational effect, regulatory mechanism, substrate selectivity, experimental evidence, 
GO term, PDB ID, cofactor, physiological pathway, drug-target potential). 
(2) Do not include the words `True/False' in the stem; hide the decisive technical point 
within the details. 
(3) Give True or False, followed by an explanation. 
Use this exact output template: Stem: …; Answer: …; Explanation: …"
\end{verbatim}\end{scriptsize}

Prompt for \texttt{Descriptive Text} generation is following:
\begin{scriptsize}\begin{verbatim}
"Based on the given annotation information of the protein, describe the given amino-acid 
sequence in one coherent paragraph that covers: 
(1) its main function and any bound cofactors/ligands, 
(2) subcellular localization and the biological process involved, 
(3) key domains or motifs, and 
(4) the specific reaction catalyzed if it is an enzyme. The description begins with A 
sentence pattern like 
`A short report on the protein with the given amino acid sequence highlights:'
or `A brief overview of the protein with the provided amino acid sequence is as follows:' 
or `A concise description of the protein with the specified amino acid sequence includes:'
or `An outline of the key aspects of the protein with the corresponding amino acid sequence 
is as follows:' 
or `A summary of the protein's main attributes with the input amino acid sequence reveals:' 
(uses similar synonymous sentences to avoid uniformity)."
\end{verbatim}\end{scriptsize}

Prompt for \texttt{Knowledge-based QA} generation is following:
\begin{scriptsize}\begin{verbatim}
"Based on the provided annotations, generate exactly 1-9 distinct, single-sentence questions 
that a researcher would naturally ask to fully interrogate this protein. Guidelines:
(1) Each question must probe a different biological dimension (expression, localization, 
mechanism, regulation, phenotype, disease, evolution, interaction, structure/properties).
(2) Keep questions concise, fluent.
(3) One per line, numbering, and the corresponding answers to these questions are concise and 
clear.
(4) The questions can be appropriately flexible and occasionally combined with some actual 
scenarios or content related to species.
The Questions and Answers are divided into two parts (wrapped with <Questions><\\Questions> 
and <Answers><\\Answers> respectively). All the information in the Q&A should be based entirely 
on the given annotations and should not be supplemented by yourself."
\end{verbatim}\end{scriptsize}

\end{document}